\pdfoutput=1

\documentclass[11pt]{article}

\usepackage[final]{coling}

\usepackage{times}
\usepackage{latexsym}
\usepackage{booktabs} 
\usepackage[T1]{fontenc}

\usepackage[utf8]{inputenc}

\usepackage{microtype}

\usepackage{inconsolata}

\usepackage{graphicx}
\usepackage{multirow}
\usepackage{makecell}

\newcommand{\tabincell}[2]{\begin{tabular}{@{}#1@{}}#2\end{tabular}}

\newcommand{\cb}[1]{\colorbox{gray!30}{#1}}

\title{Deploying Multi-task Online Server with Large Language Model}

\author{Yincen Qu\textsuperscript{\rm 1}, Chao Ma\textsuperscript{\rm 1}, Xiangying Dai\textsuperscript{\rm 1}, Hui Zhou\textsuperscript{\rm 1}, Yiting Wu\textsuperscript{\rm 1} \and Hengyue Liu\textsuperscript{\rm 2} \\
\textsuperscript{\rm 1}Trip.com Group, Shanghai, China  \\
\textsuperscript{\rm 2}Independent Researcher  \\
\normalsize{\texttt{\{yc.qu,ma\_c,xy.dai,hzhoug,ytwu\}@trip.com}} \\
\normalsize{\texttt{hengyueliu23@gmail.com}} \\
}

\begin{document}
\maketitle
\begin{abstract}
In the industry, numerous tasks are deployed online. Traditional approaches often tackle each task separately by its own network, which leads to excessive costs for developing and scaling models, especially in the context of large language models. Although multi-task methods can save costs through parameter sharing, they often struggle to outperform single-task methods in real-world applications. To tackle these challenges, we present a three-stage multi-task learning framework for large language models. It involves task filtering, followed by fine-tuning on high-resource tasks, and finally fine-tuning on all tasks. We conducted comprehensive experiments in single-task and multi-task settings. 
Our approach, exemplified on different benchmarks, demonstrates that it is able to achieve performance comparable to the single-task method while reducing up to 90.9\% of its overhead.


\end{abstract}

\section{Introduction}

In the industry, numerous natural language processing (NLP) tasks are deployed online, and all tasks are required to serve with punctuality and high accuracy. As the number of tasks increases, the demand for resources also grows. Preventing resource requirements from growing linearly with the number of tasks becomes one of the most critical challenge in cost-saving.

\begin{figure}[t]
 \includegraphics[width=0.9\linewidth]{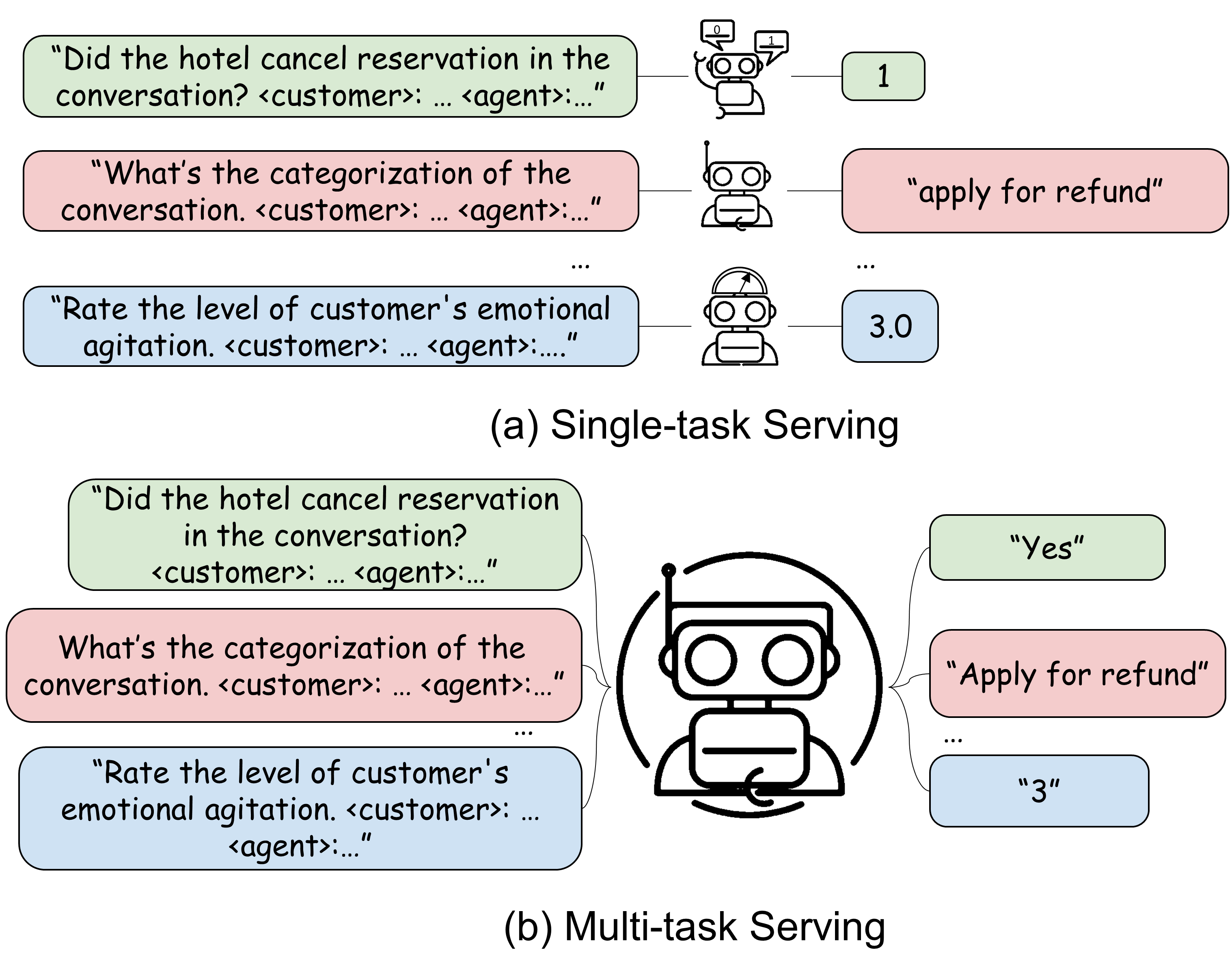} 
 \caption{Two types of online serving strategies. (a) Independent single-task models are trained and deployed for each task. (b) One multi-task model is trained and deployed for all tasks.} 
\label{figure:multitask} 
\end{figure}

Traditional approaches tackle each task separately by its own network and pipeline. This leads to excessive workloads for development and maintenance, as well as increased latency and resource usage. Moreover, in the context of large language models (LLMs), it may also lead to excessive costs for scaling up models for each task. We propose utilizing multi-task serving to deploy LLMs instead of single-task serving. \emph{single-task serving} and \emph{multi-task serving} are two types of online serving strategies, and their paradigms are shown in Figure~\ref{figure:multitask}. 
Compared to single-task serving, multi-task serving reduces deployment efforts and saves more memory due to the sharing mechanism, thus alleviating resource wastage.

However, in real-world applications, multi-task methods often struggle to match the performance of single-task methods due to the data imbalance and task heterogeneity. Data imbalance consistently leads to overfitting in low-resource tasks. This occurs because early stopping is not a feasible solution for high-resource tasks; these tasks require many more epochs to converge. Additionally, heterogeneity may result in negative transfer between tasks. Different tasks require different gradient direction in model optimization, and tasks that are too divergent may conflict in terms of gradient direction.

In this paper, we propose a three-stage framework: filtering dissimilar tasks, fine-tuning on high-resource tasks, and fine-tuning on a mixture of all tasks. The task filtering strategy prevents the negative transfer between heterogeneous tasks. The strategy of fine-tuning on high-resource tasks followed by fine-tuning on the mixture effectively effectively enables early stop by allowing different tasks to have different training epochs, thus preventing overfitting of low-resource tasks or underfitting of high-resource tasks.

Through an extensive empirical study, we find that our algorithm achieves closer performance to the single-task setting compared to other multi-task baselines. We observed that the improvement in multi-task performance mainly comes from the sampling strategy, the task filtering and domain-specific continual pre-training.


Our main contributions can be summarized as follows:

(1) We propose a framework for multi-task serving that utilizes LLMs to facilitate the multi-task method that simultaneously handles multiple tasks and achieves comparable performance of that of the single-task method. 

(2) We run a comprehensive set of experiments that suggest our scheme is practical across different benchmarks and capable of substituting for tasks trained in the single-task method. We also performed extensive experiments to gauge the importance of each of our components, such as task selection and sampling strategy.


(3) Our model was deployed to production to provide serving for a total of 11 downstream tasks. 
Compared to single-task serving, our model achieves comparable performance. We estimate that our system can reduce the total serving costs by up to 90.9\% compared to single-task serving.

\section{Related Works}
\label{sec:work}

\textbf{Multi-task Learning.}
Multi-task learning (MTL) involves training a single model on multiple tasks simultaneously.
Several studies have explored the effectiveness of MTL in various domains, such as natural language processing~\cite{hard1, mtdnn, flexible, walmart}, computer vision~\cite{hard2, longtail}. 
Recently, T5~\cite{t5}, ExT5~\cite{ext5} and Muppet~\cite{muppet} have been proposed to explore the application of Multi-Task Learning (MTL) techniques in Large Language Models (LLMs). However, they selected different checkpoints for each task without aiming to train the model to handle tasks simultaneously. Moreover, most of recent works such as FLAN~\cite{flan}, T0~\cite{t0}, and GPT-3~\cite{gpt3}, etc., focused on zero-shot or few-shot performance and neglected to compare with the full fine-tuning method for single tasks. However, we found that it is not trivial to surpass single-task full-parameter fine-tuning method.

\textbf{Data Imbalance.}
Due to the prevalence of imbalanced data distribution, data balancing has attracted increasing attention. Researchers have proposed static sampling to achieve a more balanced data distribution, which includes class-balanced sampling~\cite{classbalance}, temperature-scaled sampling~\cite{mbert}. Previous works~\cite{sampling1, sampling2} show evidence that static sampling approach yield optimal results in data rich regime (high-resources). Recently ~\citet{unimax, 2-stage} proposed to prevent model to overfit on the low-resource language in static sampling during multilingual pre-training. 
They focus on the performance of similar tasks under data imbalance, such as translation between different languages and multilingual pre-training. In our work, we integrated dissimilar tasks and explored whether data imbalance and heterogeneity could hinder multi-task performance.

\section{Preliminaries}

\begin{figure*}[htbp] 
\centering 
\includegraphics[scale=0.65]{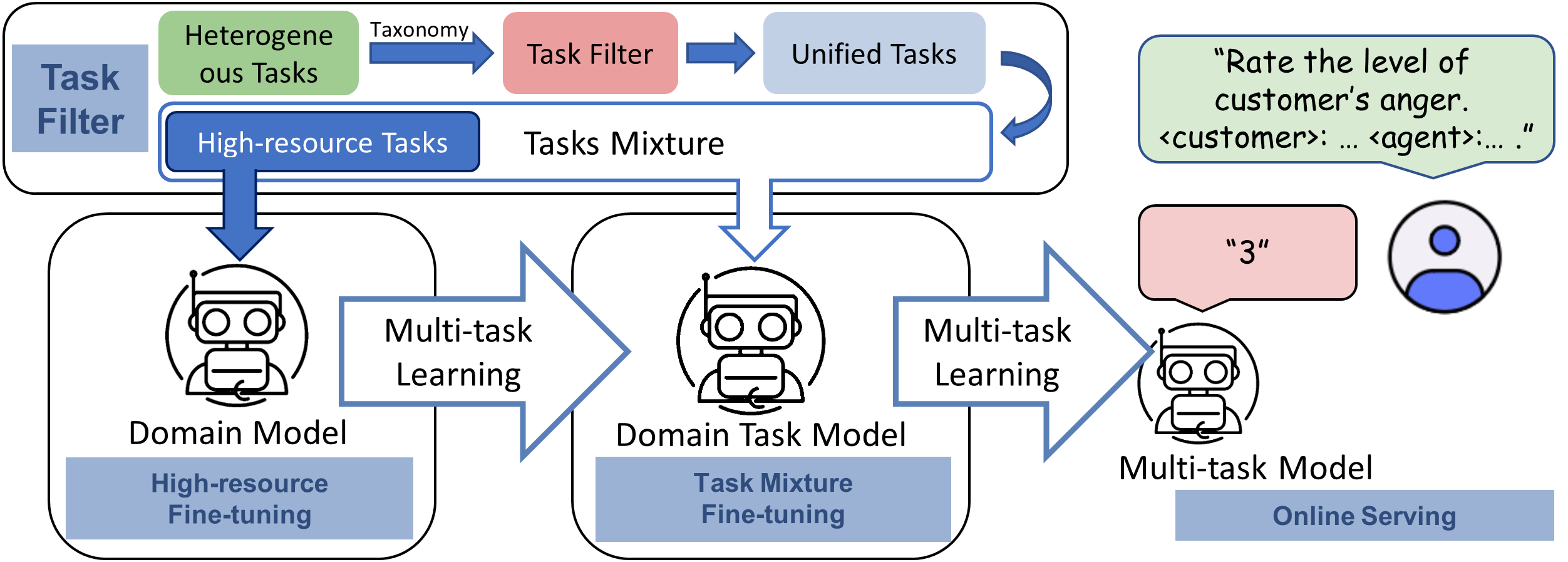} 
\caption{Pipeline of the proposed method. It starts with domain-specific continual pre-training, where the model undergoes self-supervised learning using domain-specific data. Next, we perform multi-task fine-tuning on high-resource tasks. Then, we perform multi-task fine-tuning on all tasks, enabling the model to learn from a mixture of tasks simultaneously. Finally, the multi-task model is deployed online to serve different tasks.} 
\label{figure:framework} 
\end{figure*}

\subsection{Sampling Strategies}
In this section, we present three common sampling strategies that aim to re-balance the task distribution. We will utilize these three sampling methods as baselines for subsequent experiments.

\textit{Instance-balanced sampling.} Instance-balanced sampling refers to sampling examples from each task based on the total size of each task's dataset. Specifically, the empirical distributions for different tasks are as follows.

\begin{equation}
p_l= \frac{n_l}{\sum_{ l^{\prime} \in L} n_{l^{\prime}} }
\end{equation}%
where $n_l$ is the data size of task $l$. Here data points from task $l$ will be sampled with the probability $p_l$, which is proportional to the cardinality $n_l$ of the task in the training set.

\textit{Class-balanced sampling.} Class-balanced sampling refers to sampling examples from each task with equal probability. In each batch, each example is sampled uniformly from one of the tasks used for training.

\textit{Temperature-scaled sampling.} Temperature-scaled sampling refers to re-scaling the sampling rates by a temperature $\tau$. It uses a distribution $q$ defined by exponentiating $p$.
\begin{equation}
q_l= \frac{p_l^{{1}/{\tau}}}{\sum_{l^{\prime} \in L} p_{l^{\prime}}^{1/\tau} }
\end{equation}%
When $\tau=1$, this approach is equivalent to instance-balanced sampling. As $\tau$ increases, the mixing becomes more uniform across tasks. When $\tau \to \infty$, this approach is equivalent to class-balanced sampling. In practice, commonly used values for $\tau$ are $(1.43, 2, 3.33)$~\cite{mbert, xlm, xlm-r, mt5}.

\subsection{Problem Setting}
Given a set of target tasks $L$, our framework is dedicated to find the parameters $\theta$ of a model $\mathcal F$ that can achieve comparable performance to the single-task model in as many tasks as possible. This differs slightly from the common goal of multi-task learning, which aims to achieve high average performance across all training tasks. We refer to those tasks that attain 99\% of the full fine-tuning baseline as qualified tasks and our goal is to deploy as many qualified tasks as possible with a single model. Besides, in real-world application, we have tasks of different types, each with varying amounts of training samples. Thus, we have to take heterogeneity and data imbalance into consideration.

\section{Methodology}
\label{sec:training}
Our proposed framework in Figure~\ref{figure:framework} features a pipeline that consists of three steps: 1) Task filtering; 2) High-resource task fine-tuning; 3) Tasks mixture fine-tuning. We provide a detailed breakdown of these steps below.


\subsection{Task Filter}

\subsubsection{Filtered Task}
To prevent negative transfer between different tasks, it's important to filter out inappropriate tasks. We found that generation tasks and classification tasks would hinder each others' performance in multi-task training, as evidenced in the experiment sections. The output of classification tasks is fixed, whereas the output of generation tasks is flexible. For instance, the CLUE~\cite{clue} tasks encompass single sentence classification, sentence pair classification, and machine reading comprehension. We categorize the single sentence classification and sentence pair classification as classification tasks, and machine reading comprehension as generation tasks. 

Moreover, we also investigated that whether differences in input (such as single sentences or sentence pairs) or output (such as binary classification or multi-class classification) would further impede performance. We found that the more similar the tasks are, the higher the multi-task performance can be achieved, and the greater the number of qualified tasks becomes.

\subsubsection{Unified Tasks}
\label{sec:inp}

In order to train a unified model for various tasks, we cast all of the collected tasks into a format called “text-to-text.” This format requires the model to be fed with some text for context and then generate output text for individual tasks. To indicate the specific task, we add a task-specific text prefix to the original input sequence prior to inputting it into the model. 

\subsection{Multi-task Fine-tuning}
\label{sec:finetunining}
For tasks with imbalanced data, we utilize the multi-task learning approach to balance the performance of all tasks. However, the aforementioned sampling strategies are not ideal, as they sample all tasks with a constant probability throughout the entire training process, leading to over-fitting of low-resource tasks, while high-resource tasks still require learning. 

We divide the tasks into two groups: high-resource and low-resource tasks. Since we have a variety of tasks with different training saturation steps, it is unfeasible to categorize them based on the amount of training data as in ~\citet{2-stage}. Instead, we categorize tasks based on the training saturation steps in the single-task setting. If a task achieves overfitting in fewer than 5 epochs, we refer to it as a "low-resource task." If a task achieves overfitting after more than 5 epochs, we refer to it as a "high-resource task."

For these task groups, we perform two-stage training, including high-resource task fine-tuning and tasks mixture fine-tuning.

(1) \textbf{High-resource task fine-tuning.} For high-resource tasks, we utilize the method of instance-balanced sampling to train them, given that they each have a similar amount of training data.

(2) \textbf{Tasks mixture fine-tuning.} After fine-tuning the model on high-resource tasks, we proceed to fine-tune it on the full mixture of tasks. We utilize temperature-scaled sampling and impose an artificial limit on dataset size to train all downstream tasks simultaneously. We set an artificial limit ($K$) on the dataset size to prevent over-fitting.
The adjusted distribution of different tasks is as follows.
\begin{equation}
p_l = \frac{ \textrm{min}(n_l, K)}{\sum_{ l^{\prime} \in L} \textrm{min}(n_{l^{\prime}}, K) } 
\end{equation}
\begin{equation}
q_l= \frac{p_l^{{1}/{\tau}}}{\sum_{l^{\prime} \in L} p_{l^{\prime}}^{1/\tau} }
\end{equation}




\begin{table*}[ht!]
\begin{center}
\resizebox{2.\columnwidth}{!}{%
 \begin{tabular}{ c c c c c c c c c c c} 
 \hline
 \textbf{Models} 
 & \textbf{Methods} 
 & \tabincell{c}{ \textbf{CWSC}\\(Accuracy) }
 & \tabincell{c}{ \textbf{TNEWS} \\ (Accuracy) }
 & \tabincell{c}{ \textbf{CSL} \\ (Accuracy) } 
 & \tabincell{c}{ \textbf{AFQMC } \\ (Accuracy) } 
 & \tabincell{c}{ \textbf{IFLYTEK } \\ (Accuracy) } 
 & \tabincell{c}{ \textbf{OCNLI } \\ (Accuracy) } 
 & \tabincell{c}{ \textbf{Avg.} }
 & \tabincell{c}{ \textbf{Num.} }
& \tabincell{c}{ \textbf{Overhead} }
 \\
 \hline

 \multirow{7}{*}{LLaMA} & Single-task & \cb{70.22} & \cb{58.71} & \cb{87.06} & \cb{73.98} & \cb{58.39} & \cb{79.23} & 71.26 & 6 & 100\% \\
\cline{2-11}
 & Few-shot & 65.07 & 13.82 & 62.10 & 46.80 & 14.57 & 54.47 & 42.81 & 0 & - \\
 & Instance-balanced & 68.75 & 56.20 & 85.02 & \cb{73.52} & \cb{59.13} & \cb{80.05} & 70.44 & 3 & 33.3\% \\
 & Class-balanced & 69.12 & 57.39 & 83.34 & \cb{74.05} & \cb{59.33} & \cb{80.66} & 70.64 & 3 & 33.3\% \\
 & UniMax & 68.01 & 56.55 & 84.65 & \cb{74.75} & 57.56 & \cb{82.42} & 70.65 & 2 & 50.0\% \\
 & ours & \cb{70.06} & 57.31 & \cb{87.51} & \cb{74.68} & \cb{58.79} & \cb{80.83} & 71.53 & \bf{5} & \bf{20.0\%} \\
 & ours (w/o 2-stage) & \cb{70.22} & 56.32 & \cb{87.03} & 73.03 & \cb{60.11} & \cb{81.91} & \bf{71.76} & 4 & 25.0\% \\
 \hline

 \multirow{8}{*}{Qwen} & Single-task & \cb{71.69} & \cb{60.16} & \cb{83.54} & \cb{74.12} & \cb{58.31} & \cb{86.52} & 72.39 & 6 & 100\% \\
 
 \cline{2-11}
 
 & Few-shot & 65.44 & 22.84 & 66.62 & 53.54 & 17.63 & 73.00 & 49.85 & 0 & -\\
 & Instance-balanced & 68.75 & 59.51 & \cb{82.81} & \cb{74.44} & \cb{59.56} & 82.76 & 71.30 & 3 & 33.3\% \\
 & Class-balanced & \cb{71.69} & 58.20 & \cb{85.56} & \cb{74.12} & \cb{58.54} & 80.01 & 71.35 & 4 & 25.0\%\\
 & UniMax & 70.59 & \cb{59.63} & \cb{82.74} & \cb{74.14} & \cb{59.68} & 83.37 & 71.69 & 4 & 25.0\% \\
 & ours & \cb{71.32} & \cb{59.59} & \cb{86.03} & \cb{74.56} & \cb{58.86} & 83.67 & \bf{72.33} & \bf{5} & \bf{20.0\%} \\
 & ours (w/o 2-stage) & \cb{71.32} & 58.32 & \cb{86.60} & \cb{74.18} & \cb{59.36} & 82.99 & 72.13 & 4 & 25.0\% \\
\hline
 \end{tabular}
 }
\end{center}
 \caption{Main results on 6 tasks and the average performance across them. The performance is evaluated on the development set. "Avg." refers to the macro average per-task performance of downstream tasks. "Num." refers to the amount of the qualified tasks. All metrics for tasks are multiplied by 100. \cb{Shaded} numbers indicate that they attain 99\% of the single-task fine-tuning baseline.
 } 
\label{tab:exp_task}
\end{table*}

\section{Experiments}
\label{sec:experiments}

In the following sections, we apply our proposed training method to CLUE~\cite{clue} tasks and our domain application tasks. In the CLUE experiments, we show that inappropriate sampling strategy will lead to multi-task performance degradation and different tasks taxonomies also hinder multi-task performance. In the domain-related application tasks, we scale up the number of tasks, all of which are related to the customer service field, and show that our method remains equally effective in the real-world applications.

\subsection{CLUE Tasks}

\subsubsection{Experiment Setup}
The CLUE benchmark~\cite{clue} is synthetic, consisting of six classification datasets: CWSC, TNEWS, CSL, AFQMC, IFLYTEK, and OCNLI. We provide details and references in Appendix~\ref{sec:CLUE}. For each task, we used accuracy rate as the primary evaluation metric. We reported the macro-average accuracy across all tasks within the benchmark. In the multi-task setting, we also provided the count of qualified tasks, which are defined as those achieving 99\% of the performance of their single-task counterparts. To measure the parameter and computational efficiency, we introduced a ratio: the number of qualified tasks divided by the number of models deployed. This ratio is 1 for the single-task baseline, as it deploys one model per task. For multi-task models, the ratio is calculated as 1 divided by the number of qualified tasks. This metric is labeled as "overhead" in the header of Table~\ref{tab:exp_task}.

In the experiment, we take the 7B Qwen2~\cite{qwen2} and 8B LLaMA3~\cite{LLaMA} as the base model. We present a comparative analysis of our two-stage sampling method against five benchmark approaches: few-shot prompting, single-task fine-tuning, instance-balanced sampling, class-balanced sampling, and UniMax~\cite{unimax}. In the case of few-shot prompting, we prepend five random training instances ${(q_i,a_i)}_i$ as the example to guide the model's input.

\subsubsection{Main Results}
Table~\ref{tab:exp_task} shows the experimental results on the CLUE benchmark. We observed that an inappropriate sampling strategy would hinder the multi-task performance. The few-shot method performed the worst, suggesting that it is not yet capable of directly replacing current fine-tuning methods, particularly for multi-class classification tasks. Our 2-stage sampling strategy achieved the best performance among all sampling approaches, delivering the highest number of qualified tasks. Compared to our method without the two-stage training process, the two-stage training only marginally improves average performance. However, it significantly increases the number of qualified tasks. We hypothesize that this enhancement is due to the high-resource task training helps to balance the diverse training steps across various tasks.

Moreover, we noted that LLaMA's macro-average performance on Chinese tasks is inferior to that of Qwen, likely due to insufficient training on Chinese corpora. Given that Qwen has been pre-trained on Chinese corpora, it demonstrates superior multi-task performance in Chinese. Consequently, in Section~\ref{sec:apply_experiments}, we carry out additional experiments to assess the performance of the generic model in comparison to the model that has undergone domain-specific pre-training.

\subsubsection{Taxonomy Impact}
In this section, we investigate the impact of taxonomy granularity on multi-task performance. We introduced the machine reading comprehension task CMRC into our task mixture, and trained a multi-task model with this expanded dataset. Unlike the original set of six classification tasks, CMRC, as a generation task, has a flexible output format. From the Table ~\ref{tab:clue_taxnomy1}, we found that training generation and classification tasks concurrently significantly impacts the overall performance. It is particularly notable that the performance of the classification tasks not only lags behind their single-task counterparts but also fails to match the performance of the multi-task model that was trained only on classification tasks.


To delve deeper into whether task similarity can enhance performance, we categorized the tasks into groups based on differences in input and output types: single-sentence, sentence-pair, binary classification, and multi-class classification. A more detailed presentation of the tasks and their results is provided in Appendix~\ref{sec:clue_taxonomy_impact}. From Table~\ref{tab:clue_detail_task}, we noticed that increased task similarity correlates with improved performance. However, the "overhead" metric does not decrease, as the number of models also rises. To meet our objective of cost saving, a lower overhead metric is desirable. Consequently, we decided against further subdividing these tasks into more similar categories.


\begin{table}[h]
 \centering
 \resizebox{\linewidth}{!}{

 \begin{tabular}{ l | c c c c } 
 \hline
 \textbf{Methods} 
 & \textbf{Generation}
 & \textbf{Classification} 
 & \textbf{Avg.} 
 & \textbf{Num.} \\
 \hline
 Single-task & 51.27 & 72.39 & 69.37 & 7 \\
 \hline
 instance-balanced & 47.61 & 70.64 (71.30) & 66.82 & 1 (3) \\
 class-balanced & 52.94 & 70.58 (71.35) & 68.02 & 2 (4) \\
 ours & 48.79 & 71.87 (72.33) & 68.57 & 3 (5) \\ 
\hline
 \end{tabular}
 }
 \caption{Taxonomy impact of on generation and classification CLUE tasks. The number in brackets refers to the multi-task performance trained solely with the classification tasks.
 }
\label{tab:clue_taxnomy1}
\end{table}

 

\subsection{Application Tasks}
\label{sec:apply_experiments}
In this section, we expand from a six-task setting to the setting with dozens of tasks, to verify whether task filtering and sampling methods would affect the multi-task performance. 

\subsubsection{Experiment Setup}
We tested with 17 classification tasks, which are all related to the domain of customer service. The details of these tasks are demonstrated in Appendix~\ref{sec:industry}. We also reported macro average performance, the number of qualified tasks, and the overhead metrics for each method.

We took Qwen2 7B as the base model. We provided a comparison of our method with 5 baseline methods, as in the previous section. In addition, we performed domain-specific continual pre-training on Qwen2 to obtain $\textnormal{Qwen}_d$. The details of the continual pre-training will be demonstrated in the Appendix~\ref{sec:pretraining}. We report the multi-task performance of the generic model Qwen and $\textnormal{Qwen}_d$ to further investigate whether domain pre-training can enhance multi-task performance.

\subsubsection{Application Results}
Table~\ref{tab:ind_exp} shows the experimental results on the industry benchmark. We found that when task number increases, inappropriate sampling strategy has more obvious effect on the multi-task performance. Our method outperforms other sampling baselines by consistently enhancing both the macro-average performance and the number of qualified tasks. With an overhead of only 9.1\% compared to the single-task approach, our method can potentially reduce the serving cost by up to 90.9\% relative to the single-task method.

We observed that $\textnormal{Qwen}_d$ exhibits relatively high performance compared to Qwen. Specifically, $\textnormal{Qwen}_d$ demonstrates a higher average performance than Qwen. Furthermore, any sampling method with $\textnormal{Qwen}_d$ results in a greater number of qualified tasks than with Qwen. We attribute these improvements to domain adaptation. Given the substantial disparity between customer service conversations and the general domain text corpora utilized by original LLMs, incorporating domain-specific knowledge through continuous pre-training significantly aids in downstream task performance. Moreover, the amount of required updates for each task is reduced, leading to less conflict in gradient directions when training tasks concurrently.

\begin{table}[h]
 \centering
 \resizebox{\linewidth}{!}{

 \begin{tabular}{ l c c c c } 
 \hline
 \textbf{Models} 
 & \textbf{Methods} 
 & \textbf{Avg.} 
 & \textbf{Num.} 
 & \textbf{Overhead} 
 \\
 
 \hline
 \multirow{7}{*}{Qwen} & Single-task & 88.64 & 17 & 100\% \\
 \cline{2-5}
 & Few-shot & 49.68 & 0 & -\\
 & Class-balanced & 85.34 & 5 & 20.0\%\\
 & Instance-balanced & 85.82 & 5 & 20.0\%\\
 & Unimax & 86.33 & 8 & 12.5\% \\
 & ours & 87.19 & \bf{9} & \bf{11.1\%} \\
 \hline
\multirow{7}{*}{$\textnormal{Qwen}_d$} & Single-task & 89.65 & 17 & 100\% \\
 \cline{2-5}
& Few-shot & 54.27 & 0 & -\\
& Class-balanced & 85.29 & 5 & 20.0\% \\
& Instance-balanced & 86.05 & 6 & 16.7\% \\
& Unimax & 86.91 & 8 & 12.5\%\\
& ours & 87.74 & \bf{11} & \bf{9.1\%}\\
\hline
 \end{tabular}
 }
 \caption{Main results on 17 application tasks. 
 "Avg." refers to the macro average performance. "Num." refers to the amount of the qualified tasks. 
 }
\label{tab:ind_exp}
\end{table}

\subsubsection{Taxonomy Impact}
Consistent with our previous experiment, we incorporated a generation task into our task mixture and trained them jointly with $\textnormal{Qwen}_d$. From Table~\ref{tab:ind_gc}, we found that regardless of the sampling strategy employed, both classification and generation tasks experienced a significant decline in performance compared to their single-task counterparts. This suggests that the negative impact is indeed present, likely due to the substantial differences between the tasks.

We then categorized the classification tasks into three types: binary classification, ordinal classification, and multi-class classification, and trained separate models for each category. From Table~\ref{tab:ind_bom}, we also observed that performance improved with the more granular categorization of tasks. However, since this approach required multiple models for these tasks, the overhead metric did not show improvement.

\begin{table}[h]
 \centering
 \resizebox{\linewidth}{!}{

 \begin{tabular}{ l | c c c c } 
 \hline
 \textbf{Methods} 
 & \textbf{Generation} 
 & \textbf{Classification}
 & \textbf{Avg.} 
 & \textbf{Num.} \\
 \hline
 Single-task & 57.13 & 88.64 & 86.89 & 18 \\
\hline
 class-balanced & 54.17 & 84.97 (85.29) & 83.26 & 3 (5) \\
instance-balanced & 52.58 & 85.09 (86.05) & 83.28 & 3 (6) \\
 ours & 53.69 & 85.42 (87.74) & 83.67 & 6 (11) \\
\hline
 \end{tabular}
 }
 \caption{Taxonomy impact on generation and classification application tasks. 
 }
\label{tab:ind_gc}
\end{table}

\begin{table}[h]
 \centering
 \resizebox{\linewidth}{!}{

 \begin{tabular}{ l | c c c c c c } 
 \hline
 \textbf{Methods} 
 & \textbf{Binary}
 & \textbf{Ordinal} 
 & \textbf{Multi.} 
 & \textbf{Avg.} 
 & \textbf{Num.} 
 & \textbf{Overhead} 
 \\
 \hline
 Single-task & 87.62 & 95.49 & 94.05 & 89.65 & 17 & 100\% \\
 \hline
 instance-balanced & 87.43 & 94.19 & 92.34 & 89.09 & 9 & 16.67\% \\
 class-balanced & 87.12 & 95.25 & 93.77 & 89.25 & 10 & 16.67\% \\
 ours & 87.46 & 95.21 & 93.48 & 89.43 & 10 & 14.29\% \\
 
\hline
 \end{tabular}
 }
 \caption{Taxonomy impact on binary, ordinal and multi-class classification application tasks.
 }
\label{tab:ind_bom}
\end{table}

\section{Conclusion}
In this work, we demonstrated the benefits of task filtering and two-stage multi-task training for multi-task optimization in the presence of task imbalance and heterogeneity. Through a variety of experimental setups, we show that inappropriate sampling and task selection strategies may hinder the overall multi-task performance. Our method, though straightforward, is a viable alternative to models trained with the single-task approach, potentially resulting in substantial cost savings.


\bibliography{custom}

\appendix

\section{Experiment Setting}
\label{sec:setting}

For a fair comparison, we have capped the training steps for different sampling methods at 15,000.
The hyper-parameters (e.g. learning rate, mini-batch size, etc) used in our experiments are summarized in Table~\ref{tab:parameters}.

\begin{table}
\centering
\begin{tabular}{lccc}
\hline
 
\bf{Hyper-parameter} & \bf{CLUE} & \bf{Application} \\
\hline

Learning rate & 3e-5 & 3e-5 \\
Batch Size & 1 & 1 \\
Gradient accumulation & 8 & 8 \\
Epoch (stage 1) & 1 & 1 \\
Epoch (stage 2) & 10 & 10 \\
$K$ & 20000 & 8000 \\
$\tau$ & 2 & 3.33 \\

\hline
\end{tabular}
\caption{Hyper-parameters used in our experiments.}
\label{tab:parameters}
\end{table}

\section{CLUE Benchmark}
\label{sec:CLUE}
\paragraph{Chinese Winograd Schema Challenge (CWSC).} The CWSC dataset is designed for anaphora and coreference resolution. The model is asked to determine if a pronoun and a noun phrase whithin a sentence refer to the same entity. It's a binary classification task. It mirrors similar English datasets and consists of sentences carefully selected from 36 modern Chinese literary works. Their anaphora relations are meticulously annotated by linguists, resulting in a collection of 1,838 questions.

\paragraph{TouTiao Text Classification (TNEWS).} TNEWS consists of Chinese news from TouTiao, comprising 73,360 titles in total. Each title is assigned a label among 15 different news categories, such as finance, technology and sports. The goal of this task is to predict which category the title belongs to.

\paragraph{IFLYTEK.} The IFLYTEK is a Chinese multi-class classification dataset, comprising 17,332 descriptions of mobile applications. The objective is to categorize each description into one of the 119 available categories, including but not limited to food, car rental, and education. A data filtering method akin to that employed for the TNEWS dataset has been utilized in this process.

\paragraph{Chinese Scientific Literature (CSL).} CSL dataset comprises abstracts from Chinese scientific papers and their associated keywords, sourced from various core journals across natural and social sciences. This dataset includes artificially generated keywords using the tf-idf method, which are combined with genuine keywords. The task involves identifying whether the provided keywords for a given abstract are authentic to the paper. This primarily assesses the models' capacity to determine if the keywords accurately encapsulate the content of the document.

\paragraph{Ant Financial Question Matching Corpus (AFQMC).} AFQMC originates from Ant Technology Exploration Conference (ATEC) Developer competition. It presents a binary classification challenge designed to determine if two given sentences share a similar meaning.

\paragraph{Original Chinese Natural Language Inference (OCNLI).} OCNLI is a natural language inference dataset using a similar methodology to the MNLI dataset. It consists of 56,000 inference pairs across five different categories: news, government documents, fiction, TV transcripts, and telephone transcripts. The source material for the premises is Chinese, and hypotheses were authored by university students specializing in linguistics. The level of agreement among the annotators is comparable to that of MNLI. 

\paragraph{Chinese Machine Reading Comprehension (CMRC).} CMRC is a machine reading comprehension dataset that is based on span extraction. It comprises approximately 19,071 questions, all of which are human-annotated and sourced from Wikipedia passages. Each entry in the CMRC dataset includes a context, a question, and the corresponding answer. The answers are segments of text extracted directly from the context.

\begin{table}[!htb]
\centering

\resizebox{\linewidth}{!}{
\begin{tabular}{l c c c}
\hline

 \textbf{Taxonomy}
& \textbf{Task}
& \textbf{Metrics}
& \textbf{$|{\mathcal D}|$}
\\
\hline


\multicolumn{4}{c}{\textbf{Classification}} \\

\hline
\multirow{3}{*}{Single Sentence} & CWSC & acc. & 947 \\
 &TNEWS & acc. & 49,726 \\
 &IFYTEK & acc. & 11,425 \\
\hline
 \multirow{3}{*}{Sentence Pair} & CSL & acc. & 19,836 \\
 &AFQMC & acc. & 6,564 \\
 &OCNLI & acc. & 50,437 \\

\hline
 \multicolumn{4}{c}{\textbf{Generation}} \\
\hline

Reading Comprehension & CMRC & EM. & 10,143 \\

\hline
\end{tabular}
}
\caption{ \label{tab:clue_tasks}
Examples of different tasks. $|D|$ refers to the number of training instances. 
}
\end{table}

\section{Application Tasks}

\paragraph{Reservation Cancellation (RC). }
Reservation cancellation refers to the hotel canceling a confirmed booking and not allowing guests to check-in. This is a binary classification problem where the input is a conversation, and we need to determine whether there is a booking cancellation mentioned in the conversation. Depending on the source of the input, which can be either from a phone call or an online chat, the task of reservation cancellation is considered as two separate tasks. The source of phone call is referred to as RC-A (Automatic speech recognition), while the source of online chat is referred to as RC-I (Instant messaging).

\paragraph{Unforseen Circumstances (UC).}
Unforseen circumstances refers to unforeseeable and uncontrollable circumstances that prevent guests from checking in after the hotel has confirmed a reservation. This is a binary classification problem where the input is a conversation, and we need to determine whether there is a mention of unforeseeable circumstances in the conversation. Depending on the source of the input, which can be either from a phone call or an online chat, unforseen circumstances is considered as two separate tasks. The source of phone call is referred to as UC-A (Automatic speech recognition), while the source of online chat is referred to as UC-I (Instant messaging).

\paragraph{Poaching Guests (PG).}
Poaching guests refers to persuading or forcing guests to book hotels and pay bills through alternative channels. This is a binary classification problem where the input is a conversation, and we need to determine whether there is a mention of poaching guests in the conversation. Depending on the source of the input, which can be either from a phone call or an online chat, poaching guests is considered as two separate tasks. The source of phone call is referred to as PG-A (Automatic speech recognition), while the source of online chat is referred to as PG-I (Instant messaging).


\paragraph{Insult Detection (ID).}
Insult detection is a binary classification task that determines whether a customer service representative is insulting the customer. The input for this task is the historical conversation between the customer and the customer service representative.

\paragraph{Complaint Sentiment Analysis (CSA).}
Complaint sentiment analysis refers to analyzing whether a customer is likely to post negative feedback on public platforms. The input is the customer's historical conversations, and the output is a binary classification indicating whether the conversation is likely to result in negative publicity.

\paragraph{No Room upon check-in (NR).}
No room upon check-in refers to determining whether a customer has encountered a situation where there is no available room upon their arrival at the hotel. The input is the customer's historical conversations, and the output is a binary classification. Depending on the source of the input, which can be either from a phone call or an online chat, no room upon check-in is considered as two separate tasks. The source of phone call is referred to as NR-A (Automatic speech recognition), while the source of online chat is referred to as NR-I (Instant messaging).

\paragraph{Hotel Shuttle (HS).} Hotel shuttle is a binary classification task that determines whether a hotel provides shuttle service, where the input is the conversation between the guest and the hotel. 

\paragraph{Invoice and Deposit Matters (IDM).} 
Invoice and deposit issues matters is a binary classification task. The input for this task is the conversation between the guest and the output is a binary classification indicating whether the guest requires an invoice or not.

\paragraph{Customer Service Quality Rating (CSQR).} Customer service quality rating task involves evaluating the caliber of service provided during customer interactions. For this purpose, the input data comprises historical conversations between customer service agents and their clients. The task's output is categorized into four distinct levels, numbered from 1 to 4. 

\paragraph{Scoring Extreme Emotion (SEE).} Scoring extreme emotion involves rating the level of customer agitation based on the dialogues. The resulting score ranges from 1 to 5, reflecting the intensity of their emotional state. 

\paragraph{Review Text Classification (RTC)} is a multi-label multi-class classification problem for categorizing reviews, where the input is the multi-lingual review texts and the output includes categories related to the review, such as hotel facilities, service attitude, etc. 

\paragraph{Car Services Classification (CSC).} 
Car services classification is a multi-label multi-class classification task, where the input is the historical conversation of a customer when taking a taxi, and the output is the categories of taxi-related issues mentioned by the customer.

\paragraph{Email Categorization (EC).}
Email categorization refers to classifying incoming emails based on their content. By categorizing the emails, they can be assigned to different business lines for processing. This is a multi-classification task where the input is the email content, and the output is the category of the email.



\paragraph{Conversation Summarization (CS)}. 
In the task of conversation summarization, the input consists of the historical dialogues between customer and service agents, and the goal is to produce a concise summary. 

\label{sec:industry}
\begin{table}[!htb]
\centering

\resizebox{\linewidth}{!}{
\begin{tabular}{l c c c}
\hline

\textbf{Taonomy}
& \textbf{Task}
& \textbf{Metrics}
& \textbf{$|{\mathcal D}|$}
\\
\hline


\multicolumn{4}{c}{\textbf{Classification}} \\

\hline

 \multirow{12}{*}{Binary } & RC-A & acc. & 17,059 \\
 & RC-I & acc. & 6,056\\
 
 & UC-A & acc. & 1,950 \\
 & UC-I & acc. & 8,624 \\
 
 & PG-A & acc. & 2,341\\ 
 & PG-I & acc. & 2,108 \\

 & ID & acc. & 6,884 \\

 & CSA & acc. & 5,011 \\

 & NR-A & acc. & 40,397 \\ 
 & NR-I & acc. & 19,726 \\
 
 & HS & acc. & 1,328 \\
 & IDM & acc. & 1,200 \\

\hline
\multirow{2}{*}{Ordinal} & CSQR & acc. & 2,489 \\
& SEE & acc. & 9,314 \\

\hline

\multirow{3}{*}{Multiclass} & RTC & acc. & 8,447 \\

& CSC & acc. & 8,168 \\

& EC & acc. & 6,564 \\

\hline
\multicolumn{4}{c}{\textbf{Generation}} \\

\hline


Summarization & CS & EM. & 1,822 \\
 
\hline
\end{tabular}
}
\caption{ \label{tab:ind_tasks}
Examples of different tasks. $|D|$ refers to the number of training instances. 
}
\end{table}

\begin{table*}[ht!]
\begin{center}
\resizebox{2.\columnwidth}{!}{%
 \begin{tabular}{ c c c c c c c c c c c} 
 \hline
 \textbf{Taxonomy} 
 & \textbf{Methods} 
 & \tabincell{c}{ \textbf{CWSC}\\(Accuracy) }
 & \tabincell{c}{ \textbf{TNEWS} \\ (Accuracy) }
 & \tabincell{c}{ \textbf{CSL} \\ (Accuracy) } 
 & \tabincell{c}{ \textbf{AFQMC } \\ (Accuracy) } 
 & \tabincell{c}{ \textbf{IFLYTEK } \\ (Accuracy) } 
 & \tabincell{c}{ \textbf{OCNLI } \\ (Accuracy) } 
 & \tabincell{c}{ \textbf{Avg.} }
 & \tabincell{c}{ \textbf{Num.} }
& \tabincell{c}{ \textbf{Overhead} }
 \\
 \hline

 - & Single-task & \cb{71.69} & \cb{60.16} & \cb{83.54} & \cb{74.12} & \cb{58.31} & \cb{86.52} & 72.39 & 6 & 100\% \\
\hline
 \multirow{3}{*}{SS} & Instance-balanced & 70.96 & \cb{60.42} & \cb{87.16} & \cb{74.03} & \cb{59.13} & 82.38 & 72.34 & 4 & 50.0\% \\
 & Class-balanced & \cb{73.16} & \cb{59.60} & \cb{87.40} & \cb{74.63} & \cb{59.44} & 83.64 & 72.97 & 5 & 33.3\% \\
 & ours & \cb{73.14} & \cb{60.18} & \cb{87.49} & \cb{74.32} & \cb{59.92} & 83.41 & 73.08 & 5 & 33.3\% \\

\hline

 \multirow{3}{*}{BM} & Instance-balanced & 68.01 & \cb{60.06} & \cb{87.10} & \cb{74.31} & \cb{59.29} & 84.79 & 72.25 & 4 & 50.0\% \\
 & Class-balanced & \cb{73.16} & \cb{60.10} & \cb{86.46} & 70.86 & \cb{59.60} & 84.18 & 72.39 & 4 & 50.0\% \\
 & ours & \cb{72.97} & \cb{59.91} & \cb{86.44} & \cb{73.79} & \cb{59.90} & 84.01 & 72.83 & 5 & 33.3\% \\
 \hline

 \end{tabular}
 }
\end{center}
 \caption{Results on 6 tasks with different dividing strategy. 
 } 
\label{tab:clue_detail_task}
\end{table*}

\section{CLUE Taxonomy Impact}
\label{sec:clue_taxonomy_impact}
For our CLUE dataset, we divided them into two combinations: single-sentence and sentence-pair classification, binary and multi-class classification. The single-sentence classification includes the CWSC, TNEWS, and IFLYTEK tasks, while the sentence-pairs classification includes the OCNLI, CSL, and AFQMC tasks. The binary classification includes the CWSC, CSL, and AFQMC tasks, and the multi-class classification includes the TNEWS, OCNLI, and IFLYTEK tasks. We refer to the division strategy of Single-sentence and Sentence-pairs as "SS", and the division strategy of Binary classification and Multi-class classification as "BM".

We report the detailed performance of each task in Table~\ref{tab:clue_detail_task}. As before, we also report the macro average performance, the number of qualified tasks, and the overhead. Since we have multiple models for the same benchmark, the calculation method for the "overhead" metric is slightly different from the previous one; we calculate the "overhead" by dividing 1 by the maximum number of qualified tasks per model.

\section{Continual Pre-training}
\label{sec:pretraining}
We continually pre-train the open-source foundation model on pre-processed domain-specific corpus. The following paragraphs illustrate the pre-training process, covering data sourcing, data processing, tokenization, and pre-training strategy.

\textbf{Data sourcing.}
We have collected domain-specific and general data, and mixed them together to enhance the model's general and domain-specific knowledge. Specifically, in our domain, we collect proprietary data such as customer service training materials, introductions to tourist attractions and businesses, and domain-related dialogues. Additionally, we also sample partial data from WuDaoCorpora~\citep{wudao} as general data to supplement general knowledge. This produces an approximately 150 GB collection of the pre-training corpus.

\begin{figure}[!htb]
\centering 
\includegraphics[scale=0.31]{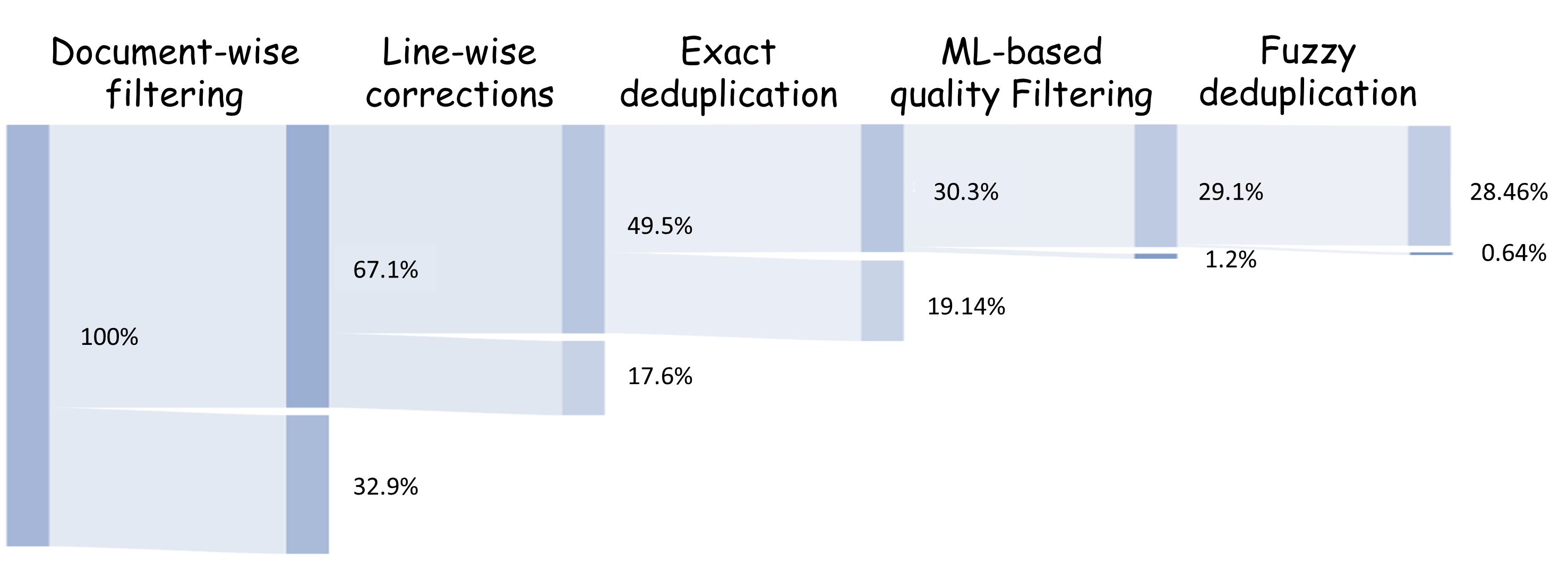}
\caption{Pipeline of data processing.} 
\label{figure:dataprocess} 
\end{figure}

\textbf{Data processing.}
We establish a comprehensive data processing pipeline to enhance pre-training data quality. This pipeline comprises four modules: document-wise filtering, line-wise corrections, exact deduplication, ML-based filtering, and fuzzy deduplication. Figure~\ref{figure:dataprocess} outlines the full data processing pipeline. After cleaning the original data, we obtain approximately 20 billion tokens of the domain-specific corpus.

\textbf{Tokenization.}
We add more domain-specific phrases as new tokens for faster training and inference. We utilize the Byte-Pair Encoding (BPE) algorithm implemented in Sentencepiece~\citep{sentencepiece} to train a domain-specific tokenizer with a vocabulary size of 13,000. We subsequently merge the domain-specific tokenizer into the original tokenizer by taking the union of their vocabularies. Specifically, the vocabulary size of the tokenizer has increased from 125,696 to 127,008. The compression rate in our domain-specific corpus has decreased from 0.6458 to 0.6104.

\textbf{Pre-training strategy.}
We utilize the self-supervised learning approach, i.e. causal language modeling, to pre-train our model on the processed corpus. Causal language models refer to models that are trained to predict the next word in a sentence based on the preceding context, capable of capturing the causal relationships between words and generating coherent text. For efficiency, we utilize Megatron~\cite{megatron} and DeepSpeed~\cite{zero} as foundational frameworks, and have integrated flash attention~\cite{flashattention}.

\end{document}